\documentclass{article}

\usepackage{microtype}
\usepackage{graphicx}
\usepackage{booktabs} 

\usepackage{caption}
\usepackage{subcaption}
\usepackage{siunitx}
\usepackage{bm}
\usepackage{amsmath}
\DeclareMathOperator{\sign}{sign}

\usepackage[hyphens]{url}
\usepackage{hyperref}

\usepackage[export]{adjustbox}

\usepackage[accepted]{mlsys2020}

\mlsystitlerunning{Larq Compute Engine}

\newcommand{\larqzoo}{\texttt{Larq Zoo}}

\begin{document}

\twocolumn[
\mlsystitle{Larq Compute Engine: Design, Benchmark, and Deploy State-of-the-Art Binarized Neural Networks}



\mlsyssetsymbol{equal}{*}

\begin{mlsysauthorlist}
\mlsysauthor{Tom Bannink*}{plumerai}
\mlsysauthor{Arash Bakhtiari*}{formerplumerai}
\mlsysauthor{Adam Hillier*}{plumerai}
\mlsysauthor{Lukas Geiger*}{plumerai}
\mlsysauthor{Tim de Bruin}{plumerai}
\mlsysauthor{Leon Overweel}{plumerai}
\mlsysauthor{Jelmer Neeven}{plumerai}
\mlsysauthor{Koen Helwegen}{plumerai}
\end{mlsysauthorlist}

\mlsysaffiliation{plumerai}{Plumerai Research. \{tom, adamh, lukas, tim, leon, jelmer, koen\}@plumerai.com.}
\mlsysaffiliation{formerplumerai}{Work done while at Plumerai Research. a.bakhtiari@tum.de}

\mlsyscorrespondingauthor{Koen Helwegen}{koen@plumerai.com}

\mlsyskeywords{Machine Learning, MLSys, TensorFlow, TFLite, TensorFlow Lite, Larq, Binarized Neural Networks, BNN, Quantization}

\vskip 0.3in

\begin{abstract}
We introduce Larq Compute Engine (LCE), a state-of-the-art Binarized Neural Network (BNN) inference engine, and use this framework to investigate several important questions about the efficiency of BNNs and to design a new leading BNN architecture. LCE provides highly optimized implementations of binary operations and accelerates binary convolutions by $8.5 - 18.5\times$ compared to their full-precision counterparts on Pixel 1 phones. LCE's integration with Larq and a sophisticated MLIR-based converter allow users to move smoothly from training to deployment. By extending TensorFlow and TensorFlow Lite, LCE supports models which combine binary and full-precision layers, and can be easily integrated into existing applications. Using LCE, we analyze the performance of existing BNN computer vision architectures and develop QuickNet, a simple, easy-to-reproduce BNN that outperforms existing binary networks in terms of latency and accuracy on ImageNet. Furthermore, we investigate the impact of full-precision shortcuts and the relationship between number of multiply-accumulate operations and model latency. We are convinced that empirical performance should drive BNN architecture design and hope this work will facilitate others to design, benchmark and deploy binary models.
\end{abstract}
]



\printAffiliationsAndNotice{\mlsysEqualContribution} 

\section{Introduction and motivation}
\label{introduction}

There are many advantages in moving deep-learning-based computer vision computation from cloud datacenters to edge devices, including
lower networking requirements, 
improved end-user privacy 
and real-time responses.
Binarized Neural Networks (BNNs) are a type of deep learning model in which a significant proportion of weights and activations are restricted to the binary values, usually $\{-1,+1\}$.
This considerably reduces model size and enables extremely efficient inference using \texttt{XOR} and \texttt{POPCOUNT} operations for binary multiplication and accumulation \cite{CourbariauxB16}.

Although BNNs have the potential to make deep learning applications radically more efficient, in practice floating point or 8-bit quantized networks still dominate deep learning models in production. We see three key reasons for this. First, training BNNs is challenging due to the gradient mismatch problem and the need for optimizing discrete weights. Second, there is a lack of integrated software tooling that can be used to rapidly develop and deploy BNNs. Third, support for BNNs on existing hardware varies, and realizing the full potential of binarization requires custom hardware. In this work we focus on the second issue.

Deep learning inference frameworks like TensorFlow Lite \cite{tensorflow_lite} have proven essential to the field of deep learning, both for research and for commercial development. These frameworks enable model development guided by direct performance measurements and quickly move from research to production.
Although several BNN inference engines have been introduced, such as BMXNet \cite{bmxnet}, DaBNN \cite{dabnn}, and Riptide \cite{riptide}, BNN research papers still tend to focus on operation counts and often lack empirical benchmarks.

In this paper, we introduce Larq Compute Engine (LCE), a state-of-the-art inference engine for Binarized Neural Networks.
We built LCE with researcher ease-of-use as a top priority, and by integrating with the TensorFlow Keras \cite{tensorflow, keras} and Larq \cite{larq} ecosystems, we provide an end-to-end pipeline for training, benchmarking, and deploying BNNs.
LCE includes a TensorFlow model graph converter and highly-optimized, hand-tuned binarized custom operators for the TensorFlow Lite runtime \cite{tensorflow_lite}.
LCE primarily targets 64-bit ARM devices, which includes all modern Android devices as well as the Raspberry Pi 3 and 4.
We show that LCE is faster than existing inference engines for both individual convolutions and complete models, through benchmarks on the Pixel 1 Android phone and Raspberry Pi Model 4B board.

We also demonstrate the power of our integrated approach by providing real-world latency benchmarks for seven major BNNs from the literature and by introducing QuickNet, a new BNN model that uses a straightforward architecture and simple single-stage training method to achieve state-of-the-art performance.
Finally, we use LCE to investigate several empirical questions about the design of BNN architectures:
(a) What are major latency bottlenecks for BNNs from the literature?
(b) What is the latency effect of different kinds of shortcut connections?
and (c) How well do MACs correlate with real-world latency?

We hope that by introducing Larq Compute Engine we provide the BNN research community with a versatile tool for designing, benchmarking, and deploying binarized models.
LCE is an actively-developed open-source project, available on GitHub at \href{https://github.com/larq/compute-engine}{\texttt{larq/compute-engine}}.

\section{Background and related work}

\subsection{Efficient network design}

While accuracy has been the primary goal of much deep learning computer vision research, model efficiency has been an increasingly important topic as DNNs have become larger and their usage as expanded. Particularly the potential for mobile end edge applications has given rise to an increasing focus on efficient network design over the past few years.

Initially, efforts were made to optimize network architectures for theoretical computational efficiency metrics.
\citet{Forrest2018} used point-wise convolutions with squeeze and expand modules to reduce the number of network parameters.
To reduce the number of operations in the network, MobileNets used depthwise separable convolutions \cite{Howard2017}, inverted residuals, and linear bottlenecks \cite{mobilenetv2}.
Grouped convolutions similarly helped reduce the number of operations with acceptable accuracy losses \cite{Zhang2018a, Huang2018}.

More recently, state-of-the-art results have been obtained through architecture searches \cite{howard2019searching, Tan2018, Chai2020}. Crucially, these searches optimize for the trade-off between accuracy and \textit{measured} quantities such as inference time or energy usage.

Besides these efforts to develop more efficient architectures, there has been a lot of work on making existing networks more efficient through pruning \cite{learningweights, Han2015} and quantization \cite{Zhu2016, Wang}. In this paper we focus on the extreme case of the latter: Binarized Neural Networks (BNNs) with 1-bit weights and activations \cite{CourbariauxB16, xnor_net}.

\subsection{Binarized neural networks}

Initial attempts of simply binarizing existing networks resulted in a large accuracy drop on all but the most simple tasks \cite{xnor_net, mobinet}.
This motivated a host of more advanced training procedures for binarized networks (e.g. \citet{Courbariaux2015, dorefa, Peters2018, Alizadeh2019, bop, real-to-binary, he2020proxybnn}), as well as changes to network architectures to make them more amenable to binarization and minimize accuracy degradation.

Some of these changes focused on more closely approximating higher bit-width networks.
One approach has been to add full-precision scaling factors to the binarized weights in order to minimize the $\ell_2$ distance from the corresponding full-precision weights \cite{xnor_net, Bulat2019b, real-to-binary}; another has been to use a combination of multiple binarized branches to approximate individual weights \cite{Lin2017} or network blocks \cite{Zhuang2018}.
\citet{Zhu2018} used an ensemble of BNNs, reducing accuracy loss at the cost of increased binary computation.
Additional residual shortcut connections for improved information flow have been found to be crucial for training more accurate BNN models~\cite{bireal_net, binary_dense_net}.

Recent BNN works have closed the gap with some of the popular higher bit-width architectures such as MobileNetv1 using custom network designs \cite{meliusnet} or by making these architectures more amenable to binarization through novel nonlinearities and training procedures \cite{reactnet}.
Neural Architecture Search has also recently been applied to BNNs \cite{Shen2019a, Bulat2020, Singh2020}.
However, while these recent works have made impressive gains in theoretical metrics, they lack evaluation on real hardware.
This pursuit of higher accuracy can lead to network designs that might look good on paper, but are hard to implement efficiently \cite{riptide}.
We therefore argue to optimize for measured quantities such as inference time, and introduce the tools required to do so.

\subsection{Existing frameworks for BNN inference}\label{sec:existing_frameworks}

There are several existing solutions for BNN inference that use \texttt{XOR} and \texttt{POPCOUNT} operations to accelerate binary multiplication and accumulation. Here we provide a brief overview of the various frameworks. We compare Larq Compute Engine to the most competitive solutions in Section \ref{sec:benchmark_results}.

\textbf{BMXNet}~\cite{bmxnet} is a BNN framework that extends MXNet~\cite{mxnet}, a general-purpose neural network training and inference library.
BMXNet implements a 2D binarized convolution operation with im2col and a binary GEMM (GEneral Matrix Multiplication) kernel, written in C++, using \texttt{XOR} operators and \texttt{\_\_builtinpopcount} compiler intrinsics.
By building on top of an existing framework BMXNet achieves broad model support without having to implement core full-precision neural network operators from scratch.
For a large batch size of 200, on Intel x86-64 CPUs with a hardware popcount instruction, BMXNet claims a 13$\times$ speed-up for their 2D binarized convolution compared to a floating point implementation with ATLAS CBLAS~\cite{atlas-cblas}.
However, the C++ binary GEMM kernel compiles to machine code that is significantly slower than what can be achieved with optimised assembly kernels.

\textbf{DaBNN}~\cite{dabnn} is a stand-alone library for BNN inference on ARM devices. Like BMXNet, DaBNN implements 2D binarized convolutions with im2col and a binary GEMM kernel, written in hand-tuned 64-bit ARM assembly.
DaBNN reports a $8 - 10\times$ speed-up for their 2D binarized convolution compared to a floating point implementation.
It is very common for BNNs to include full-precision operators, such as in the first and last layers~\cite{xnor_net}.
As DaBNN doesn't extend an existing inference framework or runtime, all supported operators must be implemented from scratch and optimised for the target platform, which limits the space of model architectures that are supported.
This approach maximises flexibility of implementation but significantly increases development cost and limits available features; for example, multi-threaded inference is not supported.

\textbf{TVM}~\cite{tvm} is a compiler stack for deploying deep learning workloads on a diverse range of hardware back-ends.
Instead of using hand-tuned optimised kernels for each operator on each target platform, the project aims to automatically generate fast kernels for running a specific model on a specific device.
Built on TVM, \textbf{Riptide}~\cite{riptide} is an end-to-end system for optimised BNN training and inference.
Models are trained with TensorFlow, and the TVM TensorFlow graph converter is modified to add support for converting binarized operators such as 2D binarized convolutions.
Riptide then extends TVM's code-generation to generate efficient kernels for these binarized operators.
Overall, Riptide reports a 4$\times$ to 12$\times$ speed-up of their BNN models compared to a floating-point implementation.
A key focus is on reducing the overhead of intermediate `glue' layers that commonly lie between pairs of binarized convolutions.
Riptide replaces these layers (weight scaling, batch normalisation, and binary re-quantisation) with a ``fused binary glue operation'' that, for example, replaces floating point multiplication by approximate scaling with a power-of-two integer shift.
The fused binary glue an effective method, though it cannot be applied when there is a residual connection between the two binarized convolutions, as is common in recent BNN literature~\cite{bireal_net, real-to-binary, meliusnet}.
The TVM compilation process means that Riptide has minimal runtime overhead, and will give good performance for a wide range of possible operators, but---as we show in Section~\ref{sec:benchmark_results}---the generated kernels do not perform as well as hand-tuned assembly kernels.

\section{Larq Compute Engine}

\begin{figure}
\includegraphics[width=\columnwidth]{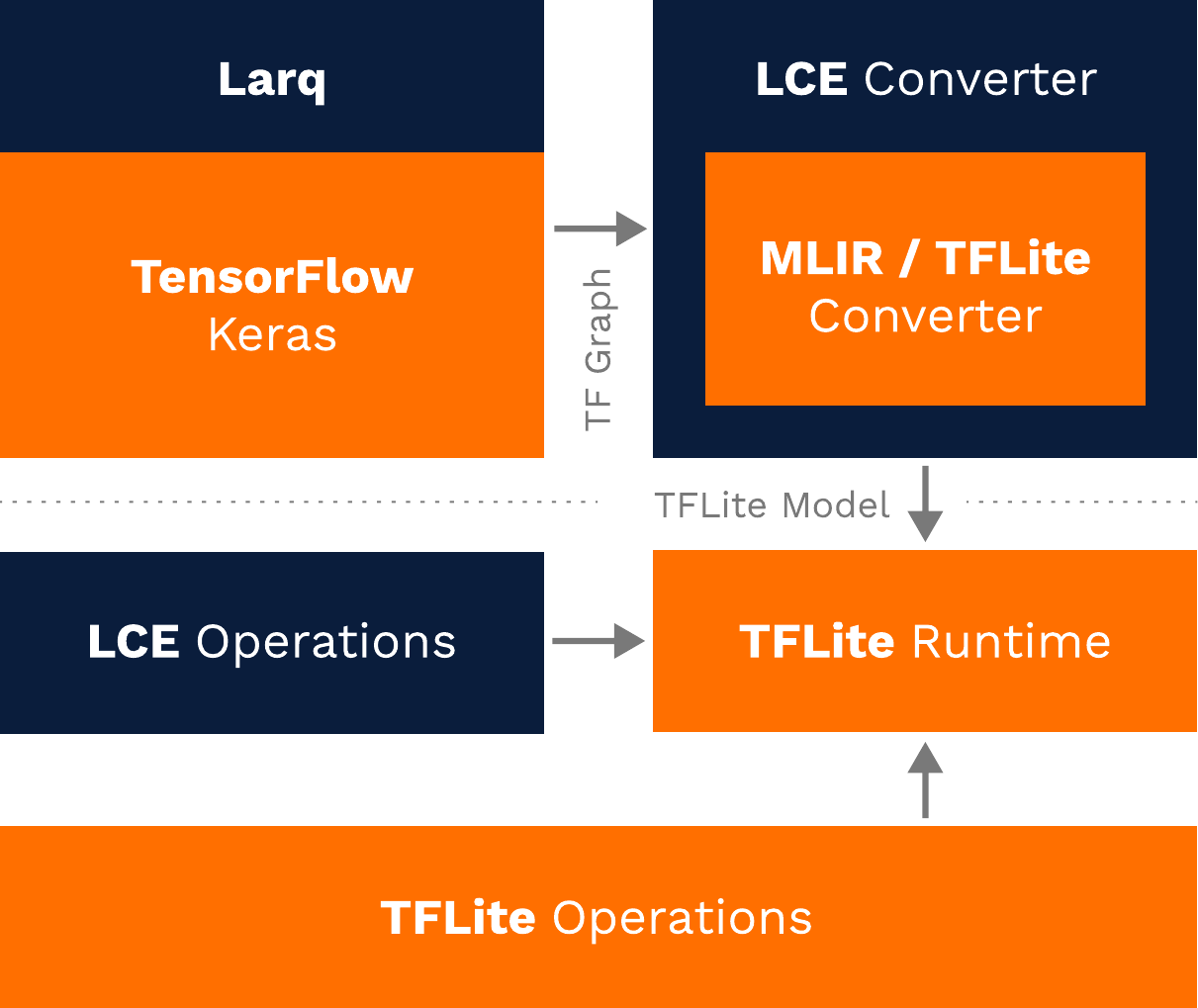}
\vskip -0.05in
\caption{Larq Compute Engine workflow from training to deployment built on-top of the TensorFlow software stack (orange) from training (top), to deployment (bottom).}\label{plot:diagram}
\end{figure}

We present Larq Compute Engine (LCE), an open-source BNN inference engine that outperforms existing BNN inference solutions. A high-level overview of the workflow from research to production is shown in Figure \ref{plot:diagram}.
LCE extends TensorFlow Lite~\cite{tensorflow_lite}, which allows us to take advantage of the existing high-performance runtime and infrastructure for model conversion, benchmarking, and deployment.
This makes it to easy adopt LCE in existing production applications.
Together with BNN training library Larq~\cite{larq} and the LCE converter, this forms an end-to-end solution for training, benchmarking and deploying BNNs. State-of-the-art models available in the \texttt{sota} submodule of Larq Zoo, the open-source library of BNN implementations, can be deployed directly with LCE.

\subsection{Conversion to inference model}\label{sec:converter}

Usability is key to enable effective use of Larq Compute Engine for researchers exploring novel architecture designs.
After building and training models with Larq~\cite{larq}, users must be able to easily convert their trained network to a TFLite model file that can be executed by our extended TFLite runtime.
We achieve this by introducing a custom converter using the MLIR~\cite{mlir} compiler infrastructure, allowing us to reuse most of the existing TFLite conversion passes \cite{tf_converters}.
This results in full support of all TFLite models and enables verification of the correctness of graph transformations applied during the conversion.

Larq, the Keras~\cite{keras}-based BNN training library, constructs a TensorFlow graph that emulates the BNN using floating point operations to approximate gradients during training. The main purpose of the converter is to transform this training graph into the TFLite model format for inference and replace the emulated binarized convolutions with truly binary, highly optimized LCE operations. 

This infrastructure allows LCE to also handle weight scaling factors as used in \citet{xnor_net} and \citet{bireal_net}, to fuse threshold-based activation functions as well as  channel-wise multipliers and biases---commonly used by batch normalization~\cite{Ioffe2015}---into the preceding binarized convolution, and to support custom padding formats for faster inference as described in Section \ref{sec:lce:ops}.
These graph transformations are crucial for efficient inference as the overhead of full-precision channel-wise operations can become significant when full-precision convolutions are replaced with binary ones.

The MLIR compiler framework makes it straightforward to add more complicated graph optimisations, too.
For example, if the output of one binarized convolution is passed through a threshold-based activation function and a batch normalization, and is then consumed by a second binarized convolution (without being used for a residual connection), there is no need to perform full-precision arithmetic or materialize the full-precision values at all.
Instead, the direct accumulator output of the first convolution can be thresholded against pre-computed values to yield the binary input to the second convolution.
The LCE converter performs these kinds of advanced optimizations automatically, without  changes to training code or instruction from the user.

The final model conversion step, after the graph optimization passes, is binary weight compression. In the Larq model graph used for training, the binary weights are stored as float values, but in the LCE model file a single bit is used for each weight value, which reduces the size of the binary weights by a factor of 32.

The converter is available to download as part of the prebuilt \texttt{larq-compute-engine} PyPI package and exposes the model conversion functionality via a single API endpoint.

\subsection{Operator implementations}
\label{sec:lce:ops}

Efficient binarized convolutions require not only a fast multiply-and-accumulation loop, but also careful design choices regarding padding, operator fusion, and more. In this section we will cover these topics and discuss some implementation details of the LCE operators.

\subsubsection*{LceQuantize}

The binarized layers in LCE expect \emph{bitpacked} input and are therefore preceded by an \texttt{LceQuantize} operator, which binarizes its input activations by extracting the sign bits\footnote{For completeness, LCE includes a \texttt{LceDequantize} operator which converts bitpacked data back into $\pm 1$-valued float data.}.
Mathematically, a $0$ valued bit represents a real value of $1.0$ while $1$ represents a real value of $-1.0$.
For optimal memory access patterns, the number of channels is padded up to a multiple of 32\footnote{Common binarized networks already have multiples of 32 channels in all their binarized layers, so in practice no padding is performed.}.
At this point, the activation tensor is 32$\times$ smaller than float input would be, and 8$\times$ smaller than 8-bit quantized input.
When the output of a binarized layer is directly used by another binarized layer (and is not required in higher precision, e.g. in a shortcut) then the first binarized layer can directly output bitpacked activations, eliminating this extra \texttt{LceQuantize} operation.

\subsubsection*{LceBConv2d}\label{sec:lce:bconv}

\begin{table*}[ht]
\newcolumntype{C}[1]{>{\centering\let\newline\\\arraybackslash\hspace{0pt}}m{#1}}
\caption{An analysis of the computational cost of performing float, 8-bit, and binary multiply and accumulate (MAC) operations using Neon SIMD instructions on the ARM Cortex-A76 CPU. In the float and 8-bit case, there exist specialized instructions that perform a fused multiplication and accumulation into 32-bit registers. Conversely, in the binary case no such fused instruction exists and so each step must be performed separately: \texttt{eor} for multiplication, \texttt{cnt} for 8-bit accumulation, and \texttt{addp} / \texttt{uadalp} for combining 8-bit results into 16-bit results. In the LCE BGEMM kernel we perform 1024 binary MACs using 24 instructions, which takes 13 cycles, or equivalently just over 78 MACs per cycle. Instruction throughput figures are sourced from the Cortex-A76 Software Optimization Guide~\cite{cortexa76_guide}. Throughput figures are theoretical sustained maximums which assume optimal instruction scheduling without CPU pipeline stalls and do not account for potential latency from loading data into registers.}\label{tab:bconv2d_instructions}
\vskip 0.15in
\centering
\begin{tabular}{p{1.5cm} C{3.8cm} C{5cm} c}
\toprule
Precision & MAC instruction sequence & Throughput (instructions / cycle) & Throughput (MACs / cycle) \\
\midrule
Float                         & \texttt{fmla} & 2 & 8 \\
\midrule
8-bit                         & \texttt{sdot} & 2 & 32 \\
\midrule
Binary  & \texttt{eor} \newline \texttt{cnt} \newline \texttt{addp} / \texttt{uadalp} & 2 \newline 1 \newline 2 / 1 & 78\\
\bottomrule
\end{tabular}
\vskip -0.15in
\end{table*}

The primary binarized operator in LCE is a 2D binarized convolution, \texttt{LceBConv2d}.
It accepts bitpacked input activations---for example, the output of a \texttt{LceQuantize} operator---and can write full-precision output or bitpacked output.
The optimized implementation of \texttt{LceBConv2d} has three stages: first, a standard \textbf{im2col} procedure is used to rearrange the input activations in memory and reduce the convolution computation to a binary matrix multiplication; second, an optimized, hand-tuned \textbf{BGEMM kernel} (Binary GEneral Matrix Multiplication) is used to perform the binary multiplication of the inputs with the weights and accumulate the results into accumulator CPU registers; and finally, an output-type-specific \textbf{output transformation} is applied which incorporates the fused channel-wise operators (see Section \ref{sec:converter}) and writes the final result to the output array.

For padded convolutions, which are very common, the im2col procedure fills the padded locations with zeros.
As per the specification of the \texttt{LceQuantize} operator, these correspond to $+1.0$ values of the original input.
We refer to this as one-padding to distinguish it from the default zero-padding in TensorFlow.
Although LCE supports zero-padded binarized convolutions, this requires an extra correction step and is therefore slower.
Larq provides the option to train binarized layers with one-padding; as we show in Section \ref{sec:quicknet}, using one-padding rather than zero-padding is not an impediment to training state-of-the-art BNNs.

The BGEMM kernel is implemented on top of the Ruy framework~\cite{ruy}, which is the GEMM library developed for use in TensorFlow Lite.
This allows us to leverage optimization techniques available in Ruy such as tiling to maximize the number of cache hits, weight packing to optimize memory access patterns, and multi-threading parallelization.
At the core of the BGEMM kernel is hand-optimized assembly code that loads data (bitpacked weights and activations) from memory into CPU registers and performs the binary multiplication and accumulation operations (MACs).
Loading binary data into registers is no different from loading float or 8-bit data; however, float and 8-bit MACs often benefit from dedicated CPU instructions which on current hardware platforms aren't available for binary MACs, which is why binarization speedups of 32$\times$ or 64$\times$ are unrealistic.
Table \ref{tab:bconv2d_instructions} shows how float, 8-bit, and binary MACs can be implemented using Neon SIMD instructions and compares the theoretical maximum MAC throughputs.
The speed of the binarized convolution also depends on how efficient the CPU cache can be used, which is where binarized layers have an advantage. For example, the weights of a binarized convolution with 256 filters of size $3\times3$ acting on 256 input channels take up 72 KiB of space which often fits entirely in the L2 cache, unlike the float or 8-bit equivalents.
In Section~\ref{sec:benchmarks} we present real world benchmarks of these implementations.

As discussed in Section \ref{sec:converter}, when writing full-precision output \texttt{LceBConv2d} supports a fused activation function and per-channel full-precision multiplier and bias.
For full-precision convolutions, the fused multiplication can be performed ``for free'' because the multiplier values can be directly folded into the convolution weights and bias.
For a binarized convolution with binary weights this is not an option, and so \texttt{LceBConv2d} has two extra inputs for per-channel full-precision values to be used as the multiplier and bias.
These fused operations are performed directly on the BGEMM accumulator values, before they are written to memory, which avoids the extra read and write that would occur without operator fusing. 
Conversely, when writing bitpacked output the BGEMM accumulator values are compared with thresholds pre-computed in the converter to decide whether each output value is a one or zero bit.

\subsubsection*{LceBMaxPool2d}

Networks that contain a full-precision MaxPool layer directly followed by a binarized convolution layer can be optimized by binarizing the activations \emph{before} the MaxPool layer instead, since $\max(\sign(X)) = \sign(\max(X))$.
The LCE converter recognizes this pattern automatically and emits the \texttt{LceBMaxPool2d} operator.
It acts on data bitpacked by the \texttt{LceQuantize} operator and simply takes the bitwise \texttt{AND} to efficiently compute the binary maxpool.

\section{Benchmarks}\label{sec:benchmarks}\label{sec:benchmark_results}

In this section we present various performance results measured using Larq Compute Engine.
The measurements were taken on a Pixel 1 phone as well as a Raspberry Pi Model 4B with a 64-bit OS (Ubuntu LTS 20.04).
The main text shows only the Pixel 1 benchmarks unless stated otherwise; the equivalent Raspberry Pi 4B numbers can be found in the appendix.
All non-binary operators use the TensorFlow Lite implementation without modifications, and all TensorFlow Lite delegates are disabled.


\subsection{The latency impact of binarizing convolutions}\label{sec:latency_impact_binarization}
\begin{figure}
\centering
\includegraphics[width=\linewidth]{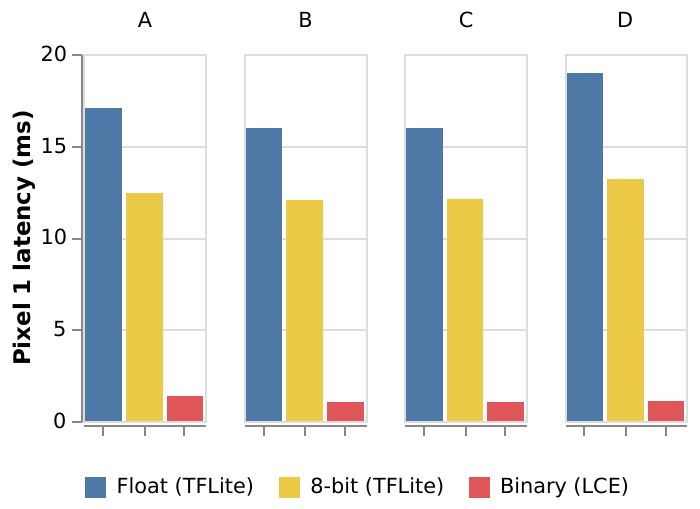}
\vskip -0.05in
\caption{The impact of binarization on latency of convolutional layers with 3$\times$3 kernels. We compare the latency of binarized convolutions to their equivalent 32-bit floating point or 8-bit integer versions for commonly used dimensions. In terms of height $\times$ width $\times$ in channels $\times$ out channels the convolutions are
(A) 56$\times$56$\times$64$\times$64;
(B) 28$\times$28$\times$128$\times$128;
(C) 14$\times$14$\times$256$\times$256;
(D) 7$\times$7$\times$256$\times$256.
Compared to floating point, we observe binary speedups of between $12\times$ for (A) and over $17\times$ for (D). Compared to 8-bit, we observe speedups of between $9\times$ and $12\times$.
}\label{plot:fl32-vs-binary-examples}
\end{figure}

We first investigate the impact of binarization on individual convolutional layers.
As an example we consider the four main convolutions that appear in ResNet18, a network architecture that has inspired numerous BNN designs.
The latencies of binarized versions of these are compared to their full-precision counterparts in Figure \ref{plot:fl32-vs-binary-examples}.
We also benchmark 8-bit quantized versions of these convolutions, as near-lossless 8-bit quantization of networks like ResNet is now commonplace.

We see a large speedup across all four layers. Binarization reduces latency by $12-17\times$ compared to the floating-point implementation, with the largest performance gains being in the layers with the most channels.

In Section~\ref{sec:lce:ops}, we explained that on the ARMv8-A platform, the CPU instructions performing the binary MACs, allowed for a theoretical throughput of 78 binary MACs per clock cycle compared to 32 8-bit MACs or 8 float MACs, if we completely ignore memory reads and other operations. These theoretical numbers would suggest a 9.75$\times$ speedup over float and a 2.43$\times$ speedup over 8-bit convolutions. Memory reads, on the other hand, would be 32$\times$ and 8$\times$ faster, respectively.
The actual speedup factors, as shown in Figure~\ref{plot:fl32-vs-binary-examples}, turn out to be higher than these theoretical MAC throughput numbers. This can be attributed to memory reads and better cache efficiency for binarized layers.




Moving beyond a handful of examples, we next investigate a large space of convolutions of different dimensions. Channels range from $\{32, 64, 96, 128, 160, 256\}$; input width and height range from $\{8, 16, 32, 64\}$, and kernel sizes are 3$\times$3 or 5$\times$5. All included convolutions preserve the dimensions of the activation tensor, i.e. they have a stride of one, use equal padding and the number of input and output channels are the same. The number of MAC operations in the investigated blocks range from roughly 0.6 million to 6.5 million, and the floating point latency on a Pixel 1 ranges from 0.01 ms to over 850 ms.

\begin{figure}
\centering
\includegraphics[width=\linewidth]{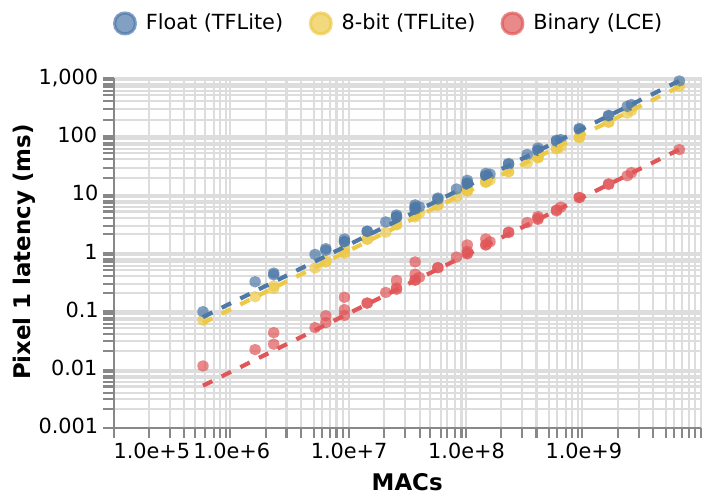}
\vskip -0.05in
\caption{The relationship between MACs and latency for a large range of convolutions in binary, int8 and 32-bit floating point. Each convolution is a dot in the figure, while the dotted lines are least-square linear regressions between the MACs and latencies. Note that we use a log-log scale. We see an approximately linear relationship between MACs and latency in each precision, especially for larger dimensions. However, we also see substantial deviations from this linear relationship even for medium-sized convolutions. The input and output activations all have the same dimensions. Channels range from $\{32, 64, 96, 128, 160, 256\}$; input width and height range from $\{8, 16, 32, 64\}$ and kernel sizes are 3 or 5.}\label{plot:conv_macs_vs_latency}
\end{figure}

The results are shown in Figure \ref{plot:conv_macs_vs_latency}.
We see  that there is an approximately linear relationship between the number of MACs and latency for all three precisions.
However, we also immediately see substantial deviations from this linear relationship.
It is clear there is not a uniform speedup even when we constrain ourselves to 2D convolutions.
These discrepancies can be caused by the overhead of bitpacking, im2col, and other operations which do not scale with MACs.

Although there is no universal speedup, we can give an approximate range of the efficiency gain one can expect from binarization on a Pixel 1. We can look at the speedup for each convolution benchmarked in Figure~\ref{plot:conv_macs_vs_latency} individually and look at the range and the mean of these speedups. Arguably, speeding up larger convolutions is more important and so we also take a weighted mean, where speedups are weighted by the full-precision latency of the block. The results are summarized in Table \ref{tab:efficiency_ratio}.

It should be noted that this speedup is highly platform-dependent and may be very different on hardware platforms with different designs, instructions, or inference frameworks.
\begin{table}
\caption{Speedup of binarized convolutions on Pixel 1 with LCE, compared to 8-bit integer or floating point precision with TensorFlow Lite. We determine this speedup for a large range of individual convolutions and provide the mean, latency-weighted mean and overall range.}\label{tab:efficiency_ratio}
\vskip 0.15in
\centering
\begin{tabular}{l c c c}
\toprule
Precision & Mean & Weighted mean & Range \\
\midrule
1 vs. 32 & 15.0$\times$ & 15.1$\times$ & 8.5--18.5$\times$  \\
1 vs. 8  & 10.8$\times$ & 11.6$\times$ & 6.1--13.4$\times$  \\
\bottomrule
\end{tabular}
\vskip -0.15in
\end{table}

\subsection{Comparison to other BNN inference frameworks}

\begin{figure}[h]

\centering
\includegraphics[width=\linewidth]{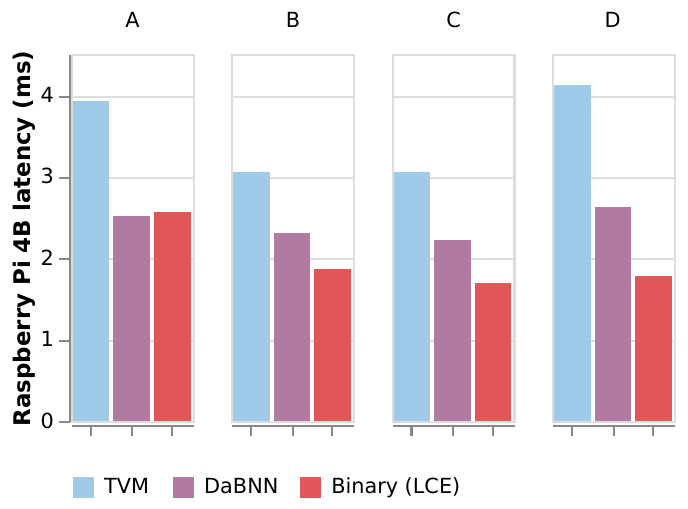}
\vskip -0.05in
\caption{Comparison of the performance of LCE versus DaBNN and TVM on representative convolutions.
The dimensions of the convolutions are the same as in Figure~\ref{plot:fl32-vs-binary-examples}.
}\label{plot:framework-benchmark}
\end{figure}

Figure~\ref{plot:framework-benchmark} shows the latencies for the same binarized convolutions as in Figure~\ref{plot:fl32-vs-binary-examples} but now measured with the different inference frameworks introduced in Section~\ref{sec:existing_frameworks}. These numbers were only measured on a Raspberry Pi 4B and not on a Pixel 1 phone because not all frameworks allowed deployment on the latter.
We have furthermore measured the overall latency of BiRealNet in these frameworks. We found it to be 119.8 ms using DaBNN, and 86.8 ms for LCE. Unfortunately the TVM latency we measured was dominated by an 830 ms initial full-precision convolution, likely due to an error somewhere causing a fallback to slower code.

\section{Designing BNNs using LCE}

In this section we leverage LCE to design accurate and efficient BNNs and to analyse performance characteristics of commonly used binarized models in literature.

\subsection{QuickNet: a simple, state-of-the-art BNN} \label{sec:quicknet}

The ability to evaluate on-device latency allows us to design novel architectures that are guaranteed to deliver on the desired accuracy-latency tradeoff.
While this opens up many possibilities, including neural architecture search and large-scale hyperparameter tuning to discover novel architectures, in the following we strive for simplicity and aim to develop an efficient network architecture that is easy to train from scratch without the need for complex multi-stage training procedures and can serve as a baseline for future research.

Our architecture follows previous work \cite{bireal_net, real-to-binary, binary_dense_net} and uses four blocks $i \in {0, 1, 2, 3}$, each consisting of $N_i$ binary 3$\times$3 convolutions with filter size $k_i$ and residual connections over each layer.
All binarized layers use one-padding (see Section~\ref{sec:lce:bconv}) and ReLU activations \cite{Glorot2011}, and are followed by a batch normalization layer \cite{Ioffe2015}.
Transition blocks between each residual section halve the spatial resolution and increase the filter count.
After the final residual block, global average pooling and a full-precision fully connected layer are used to map to the 1000 classes used by ImageNet \cite{imagenet}.

\begin{figure}
\centering
\includegraphics[width=\linewidth]{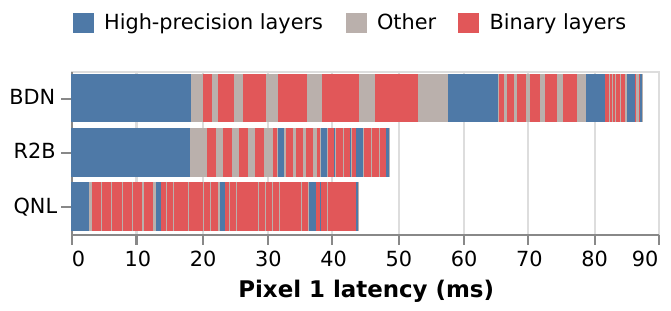}
\vskip -0.05in
\caption{Breakdown of execution latencies stacked with respect to the layer number for three models: BinaryDensent28 (BDN), RealToBinaryNet (R2B) and QuickNet Large (QNL). This clearly shows the non-negligible runtime impact of non-binary operations in BinaryDenseNet and RealToBinaryNet as well as the significant impact of the first layer in those networks. QuickNet greatly improves in both of these areas resulting in a more efficient network.}
\label{plot:model_latency_breakdown}
\end{figure}

Using the detailed operation level profiling of LCE, we can analyse similar models and clearly identify bottlenecks in the network structure.
The performance profiles of BinaryDenseNet28 and RealToBinaryNet \cite{binary_dense_net, real-to-binary} in Figure \ref{plot:model_latency_breakdown} clearly show the large impact of the first layer and other non-binary operations.

\begin{figure}
     \begin{subfigure}[b]{0.49\columnwidth}
         \centering
         \includegraphics[width=0.8\textwidth]{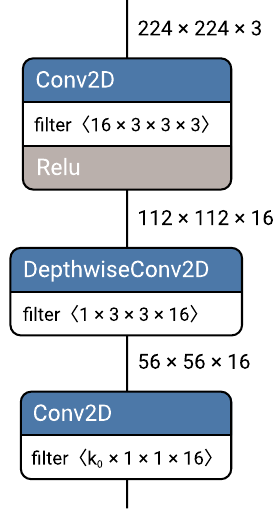}
         \caption{First convolutional block}
         \label{fig:stem}
     \end{subfigure}
     \hfill
     \begin{subfigure}[b]{0.5\columnwidth}
         \centering
         \includegraphics[width=0.8\textwidth]{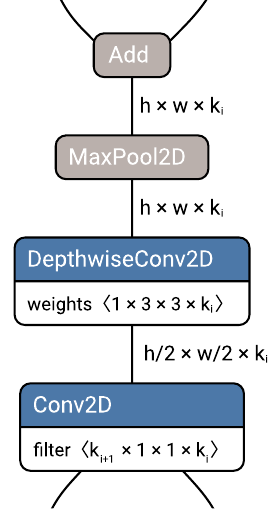}
         \caption{Transition block}
         \label{fig:transition}
     \end{subfigure}
\vskip -0.05in
\caption{Full precision blocks in QuickNet, used for spatial downsampling of (a) the input and (b) the feature map of block $i$.}
\end{figure}

To improve the efficiency of the full-precision first layer while retaining competitive accuracy, we use a small 3$\times$3 convolution with 16 filters and a depthwise separable convolution to increasing the feature size and decrease the spatial resolution from 224$\times$224 to 56$\times$56 using striding as shown in Figure~\ref{fig:stem}.
The transition block (see Figure~\ref{fig:transition}) consists of a 3$\times$3 antialiased max pooling \cite{blur_pool}---which can be efficiently implemented by a max pooling layer and a strided depthwise convolution with a fixed blurring kernel---followed by a 1$\times$1 full-precision convolutions with $k_{i + 1}$ filters to increase the feature size.

We train 3 models on the ImageNet dataset \cite{imagenet} for different latency targets and adjust the number of layers and filters according to Table~\ref{tab:quicknet_architecture}. The networks are trained from scratch for 600 epochs on 4 NVIDIA V100 GPUs with a batch size of 2048 using the Adam optimizer \cite{Kingma2014} with initial learning rate 0.01 and the straight-through estimator \cite{Hubara2016} for binary weights and stochastic gradient descent with momentum 0.9 and learning rate of 0.1 for full-precision variables. We use a linear warmup over 5 epochs for both learning rates up to their initial value and decay to zero during training using a cosine schedule.
Training images are preprocessed according to \citet{tan2019efficientnet} without AutoAugment~\cite{autoaugment} except for the largest model which slightly benefited from the additional augmentation. Table~\ref{tab:quicknet_architecture} lists the training and validation accuracies for all three models.

\begin{table}[b]
\caption{Number of layers per block $\bm{N}$ and number of filters $\bm{k}$ used in the QuickNet models and their accuracy on the ImageNet training and validation set.}
\label{tab:quicknet_architecture}
\vskip 0.14in
\centering
\small
\begin{tabular}{ccSS}
\toprule
$\bm{N}$ & $\bm{k}$ & \text{train (\%)} & \text{eval (\%)}\\
\midrule
(4, 4, \phantom{1}4, 4) & (32, \phantom{1}64, 256, 512) & 59.9 & 59.4 \\
(4, 4, \phantom{1}4, 4) & (64, 128, 256, 512) & 64.3 & 63.3 \\
(6, 8, 12, 6) & (64, 128, 256, 512) & 59.1 & 66.9\\
\bottomrule
\end{tabular}
\vskip -0.15in
\end{table}

\begin{figure}
\centering
\includegraphics[width=\columnwidth]{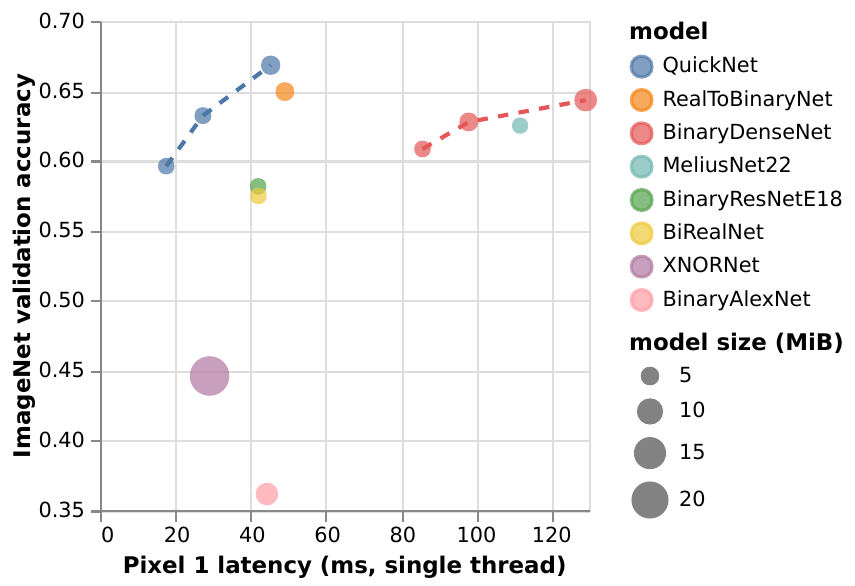}
\vskip -0.05in
\caption{Latency and accuracy for various popular BNN model architectures on the ImageNet dataset.}\label{plot:cortex_a_bnn_hero}
\end{figure}

Next we analyze the performance of QuickNet and compare against various popular binary architectures from the literature.
Reference implementations for all models discussed in this section are available in \larqzoo. We report the accuracies of the pretrained models available in \larqzoo\footnote{\url{https://docs.larq.dev/zoo/}}  which may deviate slightly from numbers reported in the original papers. Although efficiency is a key motivation behind the development of binary architectures and algorithms, most of the original papers do not measure on-device latency and instead resort to indirect measurements such as number of MACs and binary operations.

Figure~\ref{plot:cortex_a_bnn_hero} shows the accuracies and latencies of QuickNet compared to various models from previous works \cite{bireal_net, Hubara2016, binary_dense_net, meliusnet, real-to-binary, xnor_net}. We see that since the early AlexNet-based architectures, accuracies have improved substantially while memory footprints have markedly decreased. We can also observe that BiRealNet, RealToBinaryNet and QuickNet in particular have moved the pareto-front significantly forward, while other architectures such as BinaryDenseNet and MeliusNet have not fundamentally altered the landscape but rather trade higher accuracy against a worse latency.

\subsection{How do shortcuts affect BNN inference speed?}

\begin{figure}
\centering
\includegraphics[width=\columnwidth]{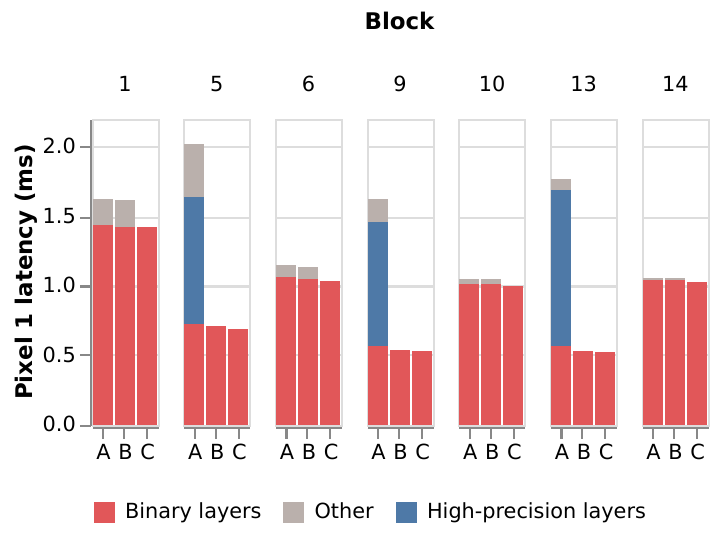}
\vskip -0.05in
\caption{We compare three versions of a binarized ResNet18: (A) with shortcuts in every block; (B) with shortcuts in the regular blocks only; and (C) with no shortcuts anywhere. Repeated layers as well as the full-precision first and last layer are not shown. We see that the latency impact of shortcuts is small for regular blocks. Unsurprisingly, for downsampling blocks which contain an additional full-precision pointwise convolution, the cost of the additional pointwise convolution is substantial.}\label{plot:shortcut_latency_impact}
\end{figure}

Since their introduction in \citet{bireal_net}, full-precision shortcuts have been pervasive in BNN architectures due to the large improvement in accuracy they provide. Such shortcuts enable the preservation of full-precision information in the forward pass and may facilitate training by carrying non-distorted gradient signals during the backward pass. They are very attractive in theoretical metrics, as they do not increase memory footprint or the number of MACs in the network.

However, they do impact the implementation of binarized networks. Whereas element-wise operations in a completely binarized architecture such as Binary AlexNet \cite{Hubara2016} can be replaced with a single binarization function, full-precision shortcuts require normal evaluation of the transformations associated to batch normalization and activation functions. Existing work such as \citet{riptide} emphasize the benefits of binarizing all intermediate activations. The question of their actual impact on latency is therefore of great practical significance.
\begin{table}[b]
\caption{The latency cost of each operator in QuickNet as a proportion of overall latency, measured on a Raspberry Pi Model 4B (single threaded). We subdivide \texttt{LceBConv2d} into the main accumulation loop (binary multiplication and accumulation) and the output transformation (integer-to-float conversion, fused activation function, and fused batch normalization).}
\vskip 0.14in
\centering
\begin{tabular}{l S}
\toprule
Operator                                    & \text{Latency (\%)} \\
\midrule
\texttt{LceQuantize}                        & 3.52  \\
\texttt{LceBConv2d} (accumulation loop)     & 53.41 \\
\texttt{LceBConv2d} (output transformation) & 3.68  \\
Full precision \texttt{Conv2D}              & 20.15 \\
Full precision \texttt{Add}                 & 9.55  \\
All other full precision                    & 9.69 \\
\bottomrule
\end{tabular}
\vskip -0.15in
\label{tab:quicknet_breakdown}
\end{table}

\begin{figure}
\centering
\includegraphics[width=0.97\columnwidth]{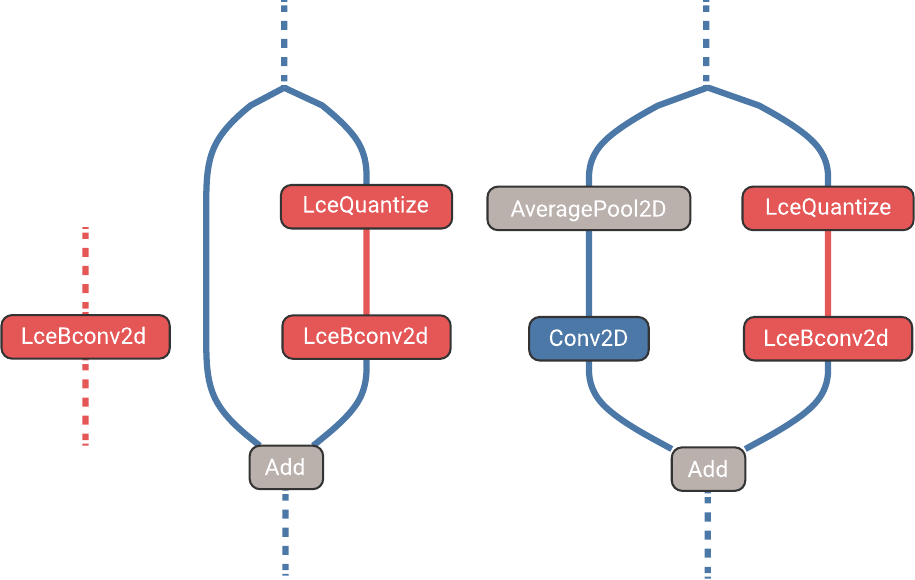}
\vskip -0.05in
\caption{Overview of the different block types that are compared in Figure~\ref{plot:shortcut_latency_impact}. Left: no shortcuts, input and output assumed to be binary. Middle: shortcut in a regular block. Right: shortcut in a downsampling layer, with a channel-doubling full-precision pointwise convolution in the shortcut.}
\label{diagram:shortcuts}
\end{figure}

To quantify the impact of full-precision shortcuts on latency, we perform latency measurements of different type of network blocks, as depicted in Figure~\ref{diagram:shortcuts}.
For the full-precision blocks, this only introduces an additional \texttt{Add} operation. For binarized operations, on the other hand, the shortcut introduces an \texttt{Add} but also forces the previous layer to write full-precision output rather than bitpacked data which means that the input activations of a subsequent binarized convolution need to be bitpacked separately, as indicated by the \texttt{LceQuantize} layer in the diagram.
Nevertheless, as we can see in Figure \ref{plot:shortcut_latency_impact}, the speed-ups remain roughly equivalent to that of binarized convolutions without shortcuts, and their absolute impact of latency is small.
Furthermore, Table \ref{tab:quicknet_breakdown} shows a breakdown of the contribution that each operator makes to overall latency of the QuickNet model, which makes it clear that the extra cost of binarized residual blocks comes from the full-precision \texttt{Add} rather than the more complex output transformation or extra bitpacking.
These results suggest that at least on this type of hardware, the use of full-precision shortcuts really does drastically improve the pareto-front for BNNs.

\subsection{Are MACs a useful proxy-metric for latency?}

MACs are still commonly used to estimate the efficiency of models despite numerous warnings stating they are an unreliable guide when searching for efficient model designs \cite{Wang, Tan2018, ma2018shufflenetv2}. This question is even more complicated in the case of BNNs because in order to come up with a scalar metric it is necessary to assume a fixed relative performance between binarized and full-precision operations. The factor 64 or values close to it is often used in the literature, based on the theoretical argument that the complexity of multiplication grows with the square of the precision \cite{bireal_net}. Some papers use the factor 58 \cite{Zhu2018, munagalastq} based on the results in \citet{xnor_net}, but this work provides no details about absolute latencies or whether any optimizations where used in the baseline benchmarks.

\begin{figure}
\centering
\includegraphics[width=\columnwidth]{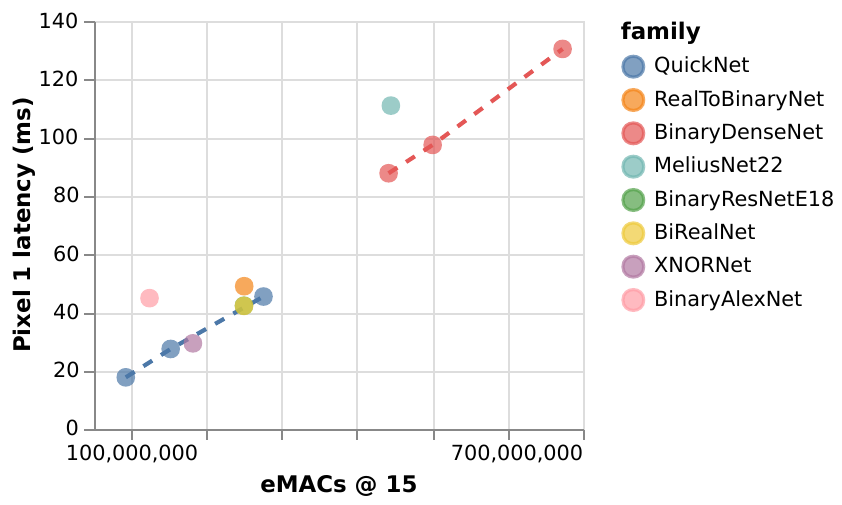}
\vskip -0.05in
\caption{The relationship between MACs and latency for the BNNs in \larqzoo. Here we assume a scaling of 15 binary MACs per full-precision MAC---the combined number is referred to as eMACs to indicate the assumed equivalence. We see MACs as a useful metric for comparing models with a similar design, but not when comparing entirely different architectures.
}\label{plot:macs_vs_latency}
\end{figure}

To get better insight into the usefullness of MACs in estimating model performance, we compare MACs and latency for the models in \larqzoo. Based on the results in section \ref{sec:latency_impact_binarization}, we assume 15 binary MACs are equivalent to one floating point MAC. The results are shown in Figure \ref{plot:macs_vs_latency}. We see that within models of the same family (e.g. QuickNets, BinaryDenseNets) MACs can be a reasonable proxy for latency. However, when comparing different model designs the relationship breaks down. For example, BinaryAlexNet is almost $2\times$ slower than models with the same number of MACs, while matching the latency of models with over $3\times$ the number of MACs.

These observations confirm that MACs have limited value when exploring new types of model designs, and cannot substitute empirical benchmarks of latency or other key performance metrics.


\section{Conclusion}\label{conclusion}

This paper introduces Larq Compute Engine, a state-of-the-art inference engine for Binarized Neural Networks. Built on top of TensorFlow and TensorFlow Lite, LCE and Larq provide an end-to-end solution for training BNNs and benchmarking them on mobile devices. The highly optimized BGEMM kernels in LCE provide speedups of $8.5\times$ to $18.5\times$ on Pixel 1 phones, while a MLIR-based converter handles the mapping from training graph to inference model, taking care of converting the emulated binary operations used during training to true binarized operations and the management of bitpacking and activation precision throughout the network. This brings software infrastructure for deploying BNNs to the same level as the infrastructure for higher precision models provided by TensorFlow and TensorFlow Lite, thus resolving one of the key obstackles to wide-scale usage of BNNs.

With LCE in place, we have been able to investigate several questions with practical importance to the development of BNNs. First, we have identified latency bottlenecks in existing network designs and shown that full-precision parts of architectures are often a major component of the overall latency of the models. Using these insights we have been able to design QuickNet---a simple, easy to reproduce binary architecture that outperforms existing binarized model in terms of accuracy and latency while using a larger fraction of binarized operations. Second, we have looked in more detail at the impact of full-precision shortcuts on latency, a topic of some controversy in the literature. We have demonstrated that although shortcuts bring some overhead in terms of latency the additional overhead is marginal and well worth the accuracy gains provided by these shortcuts. Third, we have investigated the value of MACs in designing new architectures and we have confirmed the number of MAC is a poor predictor for latency when comparing highly divergent architecture designs.

We are very excited about the future of binarized models---we see numerous opportunities and hope that LCE will facilitate further progress in the field. While QuickNet is a state-of-the-art binarized model, it is only a first step in the measurement-driven design of neural networks. On the architecture side, it has now become possible to unify the emerging field of binarized neural architecture search with the hardware-in-the-loop based approaches that have generated so much progress for full-precision models. We also note we have not focused on training methods here, and we expect QuickNet can improve further by applying more sophisticated methods such as knowledge distillation.  Above all, we hope that by providing a software framework that is high quality, easy to use and fully integrated, we will lower the barrier to experimentation with entirely novel designs and algorithms.

Finally, we want to note that hardware is a crucial component in the road towards efficient deep learning, and deployment to 64-bit ARM devices such as mobile phones is only a first step for bringing BNNs to the real world. As we discussed, most existing hardware platforms come with specialized support for full-precision or 8-bit matrix multiplication, such as vectorized MAC instructions, without providing the binarized counterpart of such operations. There is an opportunity for further large performance improvements through customized hardware for BNNs.

\section*{Acknowledgements}
We are grateful to all contributors to Larq Compute Engine and the Larq Ecosystem.

\clearpage
\bibliography{references, Plumerai}
\bibliographystyle{mlsys2020}

\appendix
\clearpage
\section{Benchmarks on Raspberry Pi 4B}

We provide the benchmark numbers on the Raspberry Pi 4B, for comparison with the numbers on a Pixel 1 phone presented in the main text.
\begin{figure}[b!]
\centering
\includegraphics[width=\linewidth]{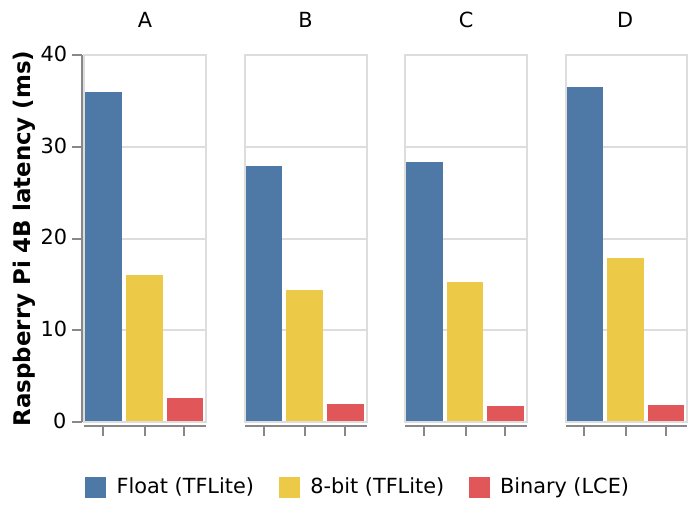}
\vskip -0.05in
\caption{The impact of binarization on latency of convolutional layers with 3$\times$3 kernel for the same convolutions as in Figure \ref{plot:fl32-vs-binary-examples}.
With respect to floating point, we observe binary speedups of between $14\times$ for (A) and over $20\times$ for (D). With respect to 8-bit, we observe speedups of between $6\times$ and $10\times$.
}\label{plot:fl32-vs-binary-examples-rpi}
\end{figure}
\begin{figure}[b]
    \centering
    \includegraphics[width=\linewidth]{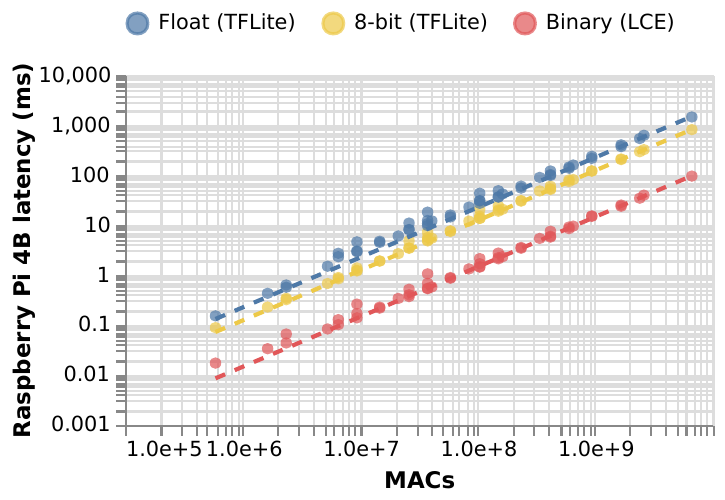}
    \vskip -0.05in
    \caption{The relationship between MACs and latency for a large range of convolutions in binary, int8 and 32-bit floating point. See also Figure \ref{plot:conv_macs_vs_latency}.}
    \label{fig:conv_macs_vs_latency_rpi}
\end{figure}

For the example convolutions, the performance for the various precisions is shown in Figure \ref{plot:fl32-vs-binary-examples-rpi}. The relationship between MACs and latency of individual convolutions is shown in Figure \ref{fig:conv_macs_vs_latency_rpi}. The associated speedups are described in Table \ref{tab:efficiency_ratio_rpi}. We see that speedups with respect to floating point convolutions is slightly better while improvement with respect to 8-bit quantized convolutions is a bit lower.

\begin{figure}[t]
    \centering
    \includegraphics[width=\linewidth]{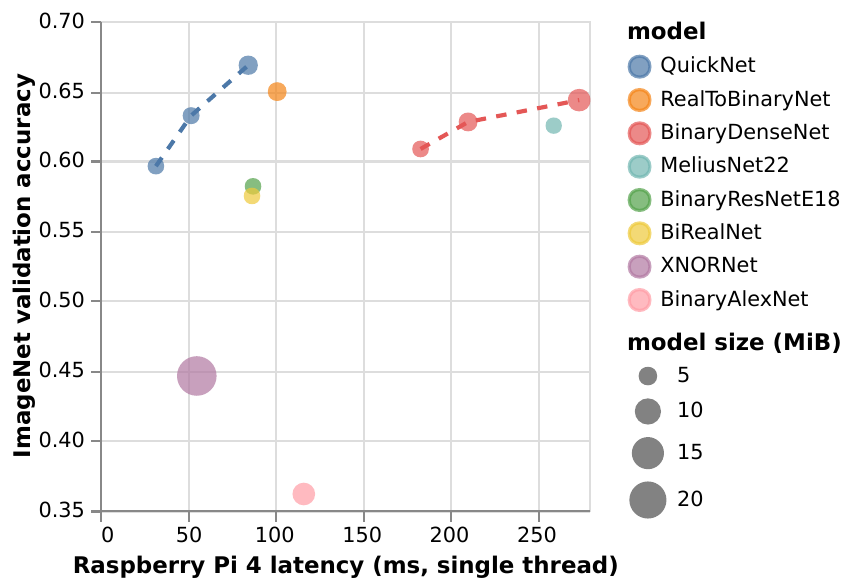}
    \vskip -0.05in
    \caption{Accuracy and latency for various popular BNN models as well as the newly introduced QuickNet on a Raspberry Pi 4B. Compare to Figure \ref{plot:cortex_a_bnn_hero}.}
    \label{fig:rpi_hero_plot}
\end{figure}

\begin{figure}
\centering
\includegraphics[width=\columnwidth]{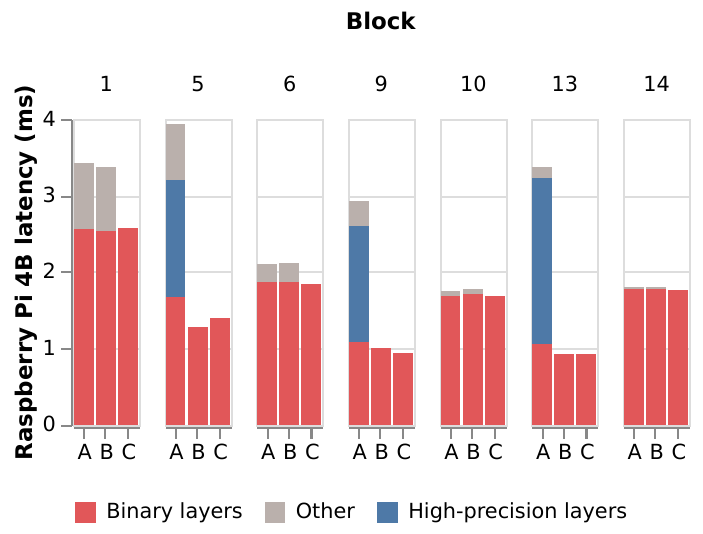}
\vskip -0.05in
\caption{Study of the impact of full-precision shortcuts on latency on a Raspberry Pi 4B. See Figure \ref{plot:shortcut_latency_impact} for details.}\label{plot:shortcut_latency_impact-rpi}
\end{figure}

The accuracy and latency of existing BNN models and QuickNet on the Raspberry Pi is shown in Figure \ref{fig:rpi_hero_plot}. The relative performance of the different models is very similar to that observed on the Pixel 1. The latency for various shortcut configurations are compared in Figure \ref{plot:shortcut_latency_impact-rpi}, and the

\begin{table}[b!]
\caption{Speedup of binarized convolutions on Raspberry Pi 4B with LCE, compared to 8-bit integer or floating point precision with TensorFlow Lite. We determine this speedup for a large range of individual convolutions and provide the mean, latency-weighted mean and overall range. Compare to Table \ref{tab:efficiency_ratio}.}\label{tab:efficiency_ratio_rpi}
\vskip 0.15in
\centering
\begin{tabular}{l c c c}
\toprule
Precision & Mean & Weighted mean & Range \\
\midrule
1 vs. 32 & 17.5$\times$ & 16.0$\times$ & 8.8--23.0$\times$  \\
1 vs. 8  & 8.3$\times$ & 8.5$\times$ & 5.1--9.6$\times$  \\
\bottomrule
\end{tabular}
\vskip -0.15in
\end{table}

\begin{figure}
\centering
\includegraphics[width=\columnwidth]{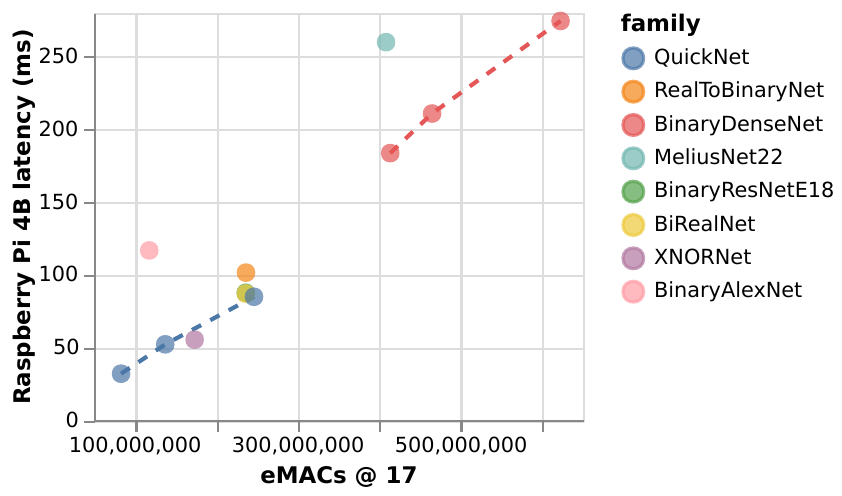}
\vskip -0.05in
\caption{The relationship between MACs and latency for the BNNs in \larqzoo. Based on Table \ref{tab:efficiency_ratio_rpi}, here we assume a scaling of 17 binary MACs per full-precision MAC - the combined number is referred to as eMACs to indicate the assumed equivalence. Compare to Figure \ref{plot:macs_vs_latency}.
}\label{plot:macs_vs_latency-rpi}
\end{figure}

\clearpage

%

\vfill

\end{document}